\relax
\documentclass[letterpaper]{llncs} 
\usepackage{times}  
\usepackage{helvet}  
\usepackage{courier}  
\usepackage{url}  
\usepackage{graphicx}
\usepackage{times}
\usepackage{graphicx}
\usepackage{latexsym}
\usepackage{color}
\usepackage{fancyhdr}
\usepackage{floatrow}
\newfloatcommand{capbtabbox}{table}[][\FBwidth]
\usepackage[T1]{fontenc}
\usepackage[utf8]{inputenc}
\usepackage[]{subcaption}

\usepackage{blindtext}

\usepackage[ruled]{algorithm2e}
\usepackage{helvet}
\usepackage{courier}
\usepackage{graphics}
\usepackage{epsf}
\usepackage{rotating}
\usepackage{times}
\usepackage{url}

\usepackage{amssymb}
\usepackage{amsfonts}
\usepackage{booktabs}
\usepackage{multirow}
\usepackage{graphicx}

\usepackage[T1]{fontenc}
\usepackage{lipsum}

\usepackage{comment}
\long\def\comment#1{}

\usepackage{comment}
\long\def\comment#1{}

\SetAlFnt{\small}
\SetAlCapFnt{\small}
\SetAlCapNameFnt{\small}
\SetAlCapHSkip{0pt}
\SetKwInOut{Parameters}{Parameters}

\definecolor{ToDoColor}{rgb}{0.1,0.2,1}
\definecolor{CommentColor}{rgb}{0.2,0.8,0.2}

\newcommand{\commentout}[1]{}

\newcommand{\glueVar}{{glue variable}}
\newcommand{\nonglueVar}{{nonglue variable}}
\newcommand{\glueVars}{{glue variables}}
\newcommand{\nonglueVars}{{nonglue variables}}

\newcommand{\MLD}{\textit{MapleLCMDist}}

\newcommand{\MLDgb}{\textit{MapleLCMDist$^{gb}$}}
\newcommand{\MLDGBShort}{\textit{MLD$^{gb}$}}

\newcommand{\MLDCBT}{{\em MapleLCMDistChronoBT}}

\newcommand{\MLDCBTgb}{{\em MapleLCMDistChronoBT$^{gb}$}}

\newcommand{\glucose}{\textit{Glucose}}

\usepackage[fleqn]{amsmath}
\usepackage{algpseudocode}

\author{Md Solimul Chowdhury \and Martin M{\"{u}}ller \textnormal\and Jia-Huai You \\
Department of Computing Science,
University of Alberta\\
Edmonton, Alberta, Canada.
\\\{mdsolimu, mmueller, jyou\}@ualberta.ca}
\title{Characterization of Glue Variables in 
CDCL SAT Solving}
\begin{document}
\pagenumbering{roman} 
\maketitle
\begin{abstract}
A state-of-the-art criterion to evaluate the importance of a given learned clause is called {\em Literal Block Distance} (LBD) score. It measures the number of distinct decision levels in a given learned clause. The lower the LBD score of a learned clause, the better is its quality. The learned clauses with LBD score of 2, called \textit{glue clauses}, are known to possess high pruning power which are never deleted from the clause databases of the modern CDCL SAT solvers. In this work, we relate glue clauses to decision variables. We call the variables that appeared in at least one glue clause up to the current search state \glueVars{}.  We first show experimentally, by running the state-of-the-art CDCL SAT solver \MLD{} on 
benchmarks from SAT Competition-2017 and 2018, that branching decisions with glue variables are categorically more inference and conflict efficient than nonglue variables. Based on this observation, we develop a structure aware CDCL variable bumping scheme, which bumps the activity score of a glue variable based on its appearance count in the glue clauses that are learned so far by the search. Empirical evaluation shows effectiveness of the new method over the main track instances from SAT Competition 2017 and 2018. 

\end{abstract}   

\section{Introduction}
SAT is known to be NP-complete \cite{Cook71}. Despite the hardness, modern CDCL SAT solvers routinely solve large real world instances with thousands of variables and millions of clauses with surprising efficiency. This is the result of elegant mashup of its key components, such as preprocessing/inprocessing, robust branching heuristics, efficient restart policies, intelligent conflict analysis, and effective \textit{clause learning} \cite{HandbookSAT}. 

Clause learning facilitates search by pruning the search space. However, since a clause learning CDCL solver learns clauses at a very high rate and a large amount of learned clauses may reduce the overall performance, \textit{clause database management} has become another key component of modern CDCL SAT solving, which periodically reduces the learned clause database by keeping only the most relevant ones \cite{Oh2016PhD}. 

Before the CDCL SAT solver \glucose{} \cite{AudemardS09}, size and recent activity of the learned clauses were the dominant criteria for determining relevance of learned clauses \cite{MiniSATPaper}. \glucose{} proposed a new measure named {\em Literal Block Distance} (LBD) score, which measures the number of distinct decision levels in a learned clause. 
\glucose{} (as well as some other modern CDCL SAT solvers) permanently stores a special type of learned clauses, called \textit{glue clauses}, which have LBD score of 2. A glue clause connects, under the current partial assignment,
a literal from the current decision level with the block of literals propagated in a previous decision level. As a glue clause connects a block of closely related variable, a relatively small number of decisions are needed to activate (i.e., making it unit) that glue clause in the future. Thus learning of a glue clause may cause relatively higher number of propagations and faster generation of conflicts within fewer numbers of decisions.  Simply put, glue clauses have more potential to reduce the search space more quickly than other learned clauses with higher LBD scores.  


Inspired by the intuitive characteristics of glue clauses, we ask the following question: Can glue clauses be used to help re-rank decision variables to improve search efficiency? 
To address this question, we consider two types of variables,  \glueVars{}, the ones that have appeared in at least one glue clause up to the current search state and \nonglueVars{} that have not. 
The main contributions of this paper are: 
\begin{itemize}
\item We conduct an experiment using the instances from the main track of SAT Competition 2017 and 2018 (abbreviated as SAT-2017 and SAT-2018, respectively) with the CDCL SAT solver \MLD{}.
First, our experiment shows that decisions with glue variables are categorically more inference and conflict efficient than nonglue variables. Secondly, comparing with nonglue variables, glue variables are picked up by the branching heuristics of \MLD{} disproportionately more often. 

\item The above results motivated us to design a structure-aware variable score bumping method named \textit{Glue Bumping} (GB), based on the notion of \textit{glue centrality} of glue variables, which dynamically measures the occurrence count of a \glueVar{} relative to the occurrence count of all the \glueVars{} discovered so far by the search. This differs from the score bumping in \glucose{} \cite{AudemardS09}, which increases the scores of variables in a learned clause that were propagated by a glue clause. Our idea is to bump the activity score of a \glueVar{} based on its glue centrality score, encouraging the CDCL branching heuristics to branch more often on recently active high centrality \glueVars{}.  We implement the GB method on top of 
\MLD{} (winner of SAT-2017)  and \MLDCBT{} (winner of SAT-2018).
Our GB extensions achieve lower PAR-2 scores than the baselines over the instances from SAT-2017 and SAT-2018, which shows the effectiveness of the new method. 
\end{itemize}
\section{Notations}
 Let $\mathcal{F}$ be a SAT formula and $vars_{\mathcal{F}}$  the set of variables in $\mathcal{F}$. Suppose a CDCL solver $M$ is solving $\mathcal{F}$ and $s$ is the current search state. At $s$, $M$ has taken $\#d>0$ decisions, generated $\#p\ge 0$ propagations, $\#c \ge 0$ conflicts, and learned $\#g \ge 0$ glue clauses. We denote the set of these $g$ glue clauses by $G$. We define {\em Propagation Rate} (PR), {\em Learning Rate} (LR) over $\#d$ decisions as $\frac{\#p}{\#d}$ and  $\frac{\#c}{\#d}$, respectively.   
 
 Let $vars_G$ be the set of \glueVars{} appearing in $G$ and $vars_{NG}=vars_{\mathcal{F}} \setminus vars_{G}$ be the set of \nonglueVars{}. We define \textit{Glue Level}, denoted $gl(v_g)$, of a \glueVar{} $v_g$ as the number of glue clauses in $G$ in which $v_g$ appears. 
 \textit{Glue Centrality} for the \glueVar{} $v_g$, denoted by $gc_{v_g}$, is the ratio of its glue level and the combined glue level of all glue variables in $vars_G$, i.e.,  $gc(v_g)= \frac{gl(v_g)}{\forall_{v_g' \in vars_{G}} \sum gl(v_g')}$. 
 

A \textit{Glue Decision} is the branching decision that selects a glue variable to branch on. Similarly, a \textit{NonGlue Decision} is the branching decision that selects a \nonglueVar{} to branch on. Suppose $M$ has taken $\#gd\le \#d$ glue decisions and $\#ngd \le \#d$ nonglue decisions until $s$. 
We define \textit{Glue variable Fraction (GF)} (resp. \textit{NonGlue variable Fraction (NGF)} )  as $\frac{\vert vars_G \vert}{\vert vars(\mathcal{F}) \vert}$ (resp. $\frac{\vert vars_{NG} \vert}{\vert vars(\mathcal{F}) \vert} $), which denotes the fraction of variables in $\mathcal{F}$, which are glue (resp. nonglue ) variables until $s$. 

\section{Experiment with Glue Variables}

In this section, we  report experimental results regarding the role of glue variables in a CDCL solving process. The CDCL solver in this experiment is \MLD{}, which uses a combination of three CDCL heuristics VSIDS \cite{Chaff}, LRB \cite{LiangGPCSAT16}, and Dist \cite{dist}. 

We run all 750 instances used in the main track of SAT-2017 (350 instances) and 2018 (400 instances) with 5000 seconds timeout limit per instance. We slightly modified \MLD{} to collect the following statistics for each instance: (i) the numbers of glue and nonglue decisions, (ii) PR, LR, average LBD for both glue and nonglue decisions, and (iii) GF/NGF.  All the experiments are run on a Linux workstation with 64 GigaBytes RAM and processor clock speed of 1.2 GHZ.


\vspace{-.1in}
\begin{table}[h!]
\caption{Comparison of average of PR, LR and avg. LBD of glue and nonglue decisions.} 
\label{TabBaseComparision}
\resizebox{0.7\textwidth}{!}{%
	{\small
\begin{tabular}{|c|c|c|c|c|c|c|c|}
\hline
\multirow{2}{*}{\textbf{Type}} & \multirow{2}{*}{\textbf{\#Instance}} & \multicolumn{3}{c|}{\textbf{Glue Decisions Average}} & \multicolumn{3}{c|}{\textbf{NonGlue Decisions Average}} \\ \cline{3-8} 
                               &                                      & PR                & LR             & aLBD            & PR                 & LR              & aLBD             \\ \hline
Satisfiable                    & 235                                  & \textbf{309.58}   & \textbf{0.47}  & \textbf{31.76}  & 61.42              & 0.19            & 40.55            \\ \hline
Unsatisfiable                  & 207                                  & \textbf{417.49}   & \textbf{0.59}  & \textbf{12.82}  & 65.84             & 0.27            & 30.10            \\ \hline
Unsolved                       & 308                                  & \textbf{385.08}   & \textbf{0.52}  & \textbf{24.23}  & 153.39             & 0.37            & 34.09            \\ \hline
Overall                        & 750                                  & \textbf{371.21}   & \textbf{0.52}  & \textbf{23.29}  & 100.46            & 0.29            & 35.02          \\ \hline
\end{tabular}}}
\end{table}

\noindent{\bf Inference and Conflict Generation Power of Glue Variables:}
Table \ref{TabBaseComparision} shows a comparison of average PR, LR and average LBD for glue and nonglue decisions, grouped by satisfiable, unsatisfiable and unsolved instances.  Compared to the nonglue decisions, on average, \MLD{} achieves significantly higher PR (overall almost 4 times higher), LR (overall, almost twice as high) and lower average LBD (overall, almost 1.5 times lower) with the glue decisions.  

\begin{figure}[h!]
\begin{center}
\caption{Comparison of LR, PR and average LBD scores. In the left, middle and right plot, instances are sorted by PR, LR and average LBD scores of glue decisions, respectively. } 
\label{GRConfEffvsNGDConfEff}
\centering
\includegraphics[width=\textwidth, height=0.5\textwidth]{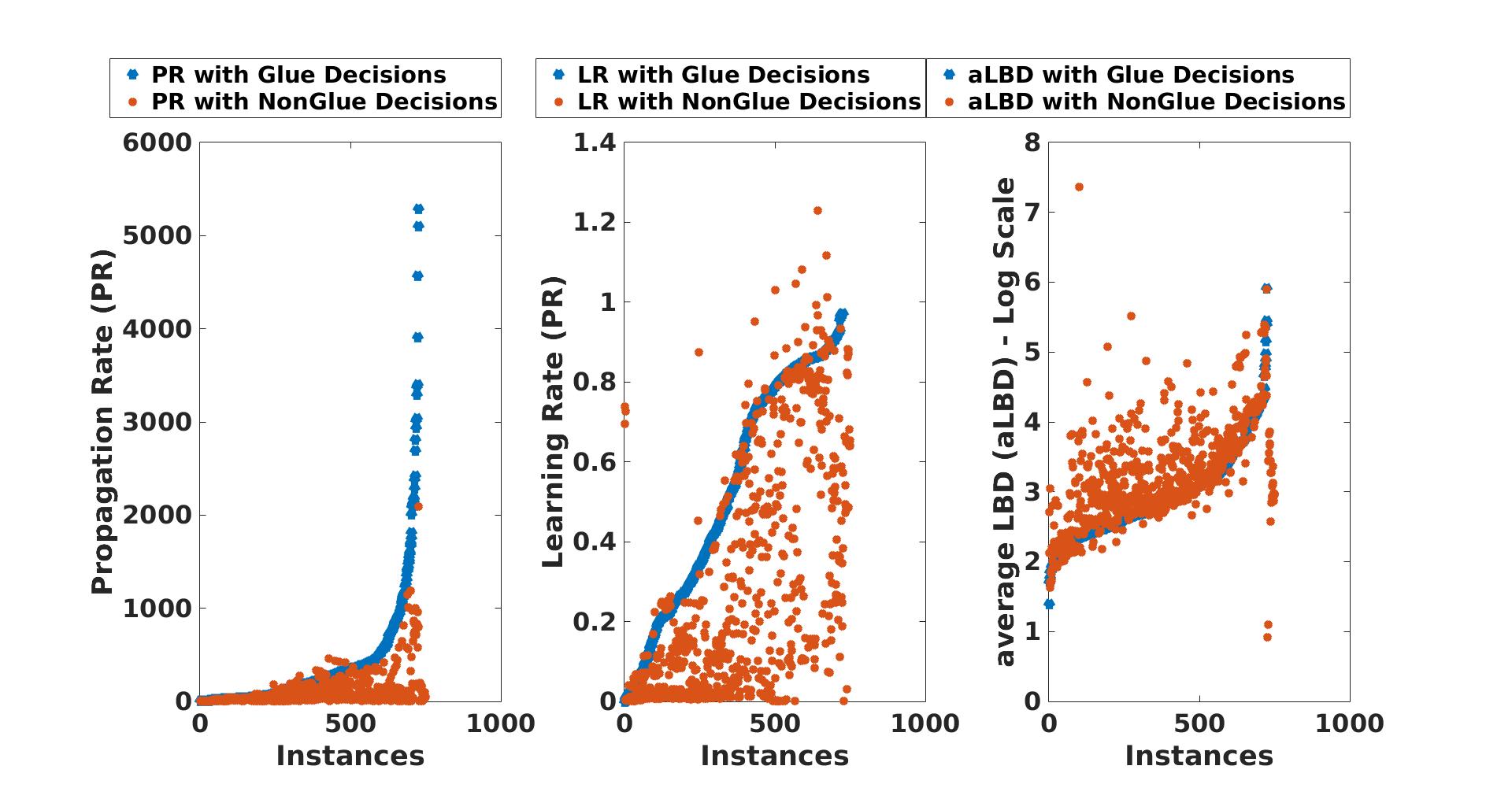}
\end{center}
\end{figure}

\begin{figure}[h!]
\begin{center}
\label{GRConfEffvsNGDConfEff}
\caption{Disproportionate selection of \glueVars{} wrt. their relative pool size.}
\centering
\includegraphics[width=\textwidth, height=0.5\textwidth]{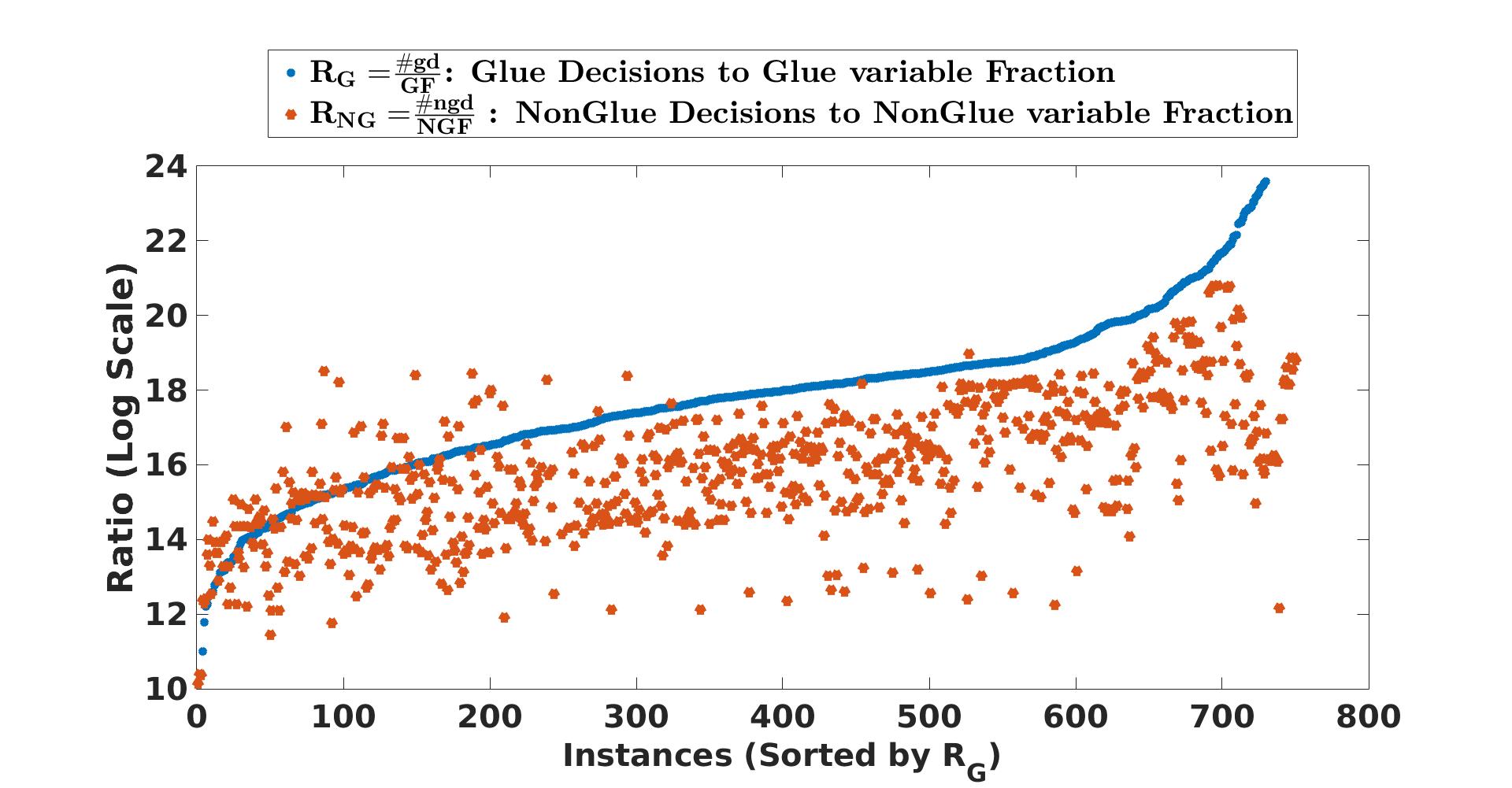}
\end{center}
\end{figure}

Figure 1 shows the PR (left), LR (middle) and average LBD (right, in log scale) values  for both glue and nonglue decisions. Clearly,  these three plots in this figure accurately reflects the average values of PR, LR and average LBD in Table \ref{TabBaseComparision}. For almost all the instances glue decisions achieve higher PR and LR compared to nonglue decisions. Compared to the nonglue decisions, the average LBD score of the clauses which are learned from the conflicts generated with glue decisions is lower for almost all the instances. Thus glue decisions are categorically more inference and conflict efficient than nonglue decisions.


\medskip
\noindent{\bf Selection Bias of Glue Variables:}
A question arises as whether conflict guided CDCL branching heuristics exhibit any bias towards 
\glueVars{} over \nonglueVars{}.  

The average value of the number of glue decisions ($\#gd$) and the number of  nonglue decisions ($\#ngd$) over the 750 instances are 26857888.34 (> 26.8 M) and 32007216.51 (>32 M), respectively.  The average GF (\glueVar{} Fraction) and NGF (\nonglueVar{ Fraction}) values over these 750  instances are 0.22 and 0.78, respectively. Thus, on average, from a $\frac{0.78-0.22}{0.22} *100=254.54\%$ bigger pool, on average, \nonglueVars{} are selected only $\frac{32007216.51-26857888.34}{26857888.34} *100=19.17\%$ more often than \glueVars{}. Hence, CDCL heuristics exhibits a clear bias towards the selection of \glueVars{} over \nonglueVars{}.

Figure 2 compares two ratios, $R_G = \frac{\#gd}{GF}$ and $R_{NG}=\frac{\#ngd}{NGF}$ in log scale. $R_G$ (resp. $R_{NG}$) measures the number of glue decisions (resp. nonglue decisions) w.r.t. glue (resp. nonglue) variable pool size for the 750 instances. It clearly shows that glue decisions relative to \glueVars{} pool fractions are higher than nonglue decisions relative to the \nonglueVars{} pool fractions for most of the instances.

To summarize, decisions with glue variables are categorically more conflict efficient than decisions with nonglue variables. As a result, CDCL branching heuristics exhibits clear bias towards glue variables over nonglue variables.   

\section{Activity Score Bumping for Glue Variables}
From the above analysis, it is clear that decisions with glue variables are more conflict efficient than with nonglue variables.  An interesting question is how we can exploit this empirical characteristic for more efficient CDCL SAT solving. 

In this section, we present a score bumping method, called {\em  Glue Bump} (GB), for bumping the activity scores of glue variables on top of the standard activity-based CDCL branching heuristics. The design of this combination is to give  more preference towards the selection of recently active high centrality glue variables. 
\paragraph{The GB Method} The GB method modifies a CDCL SAT solver $M$ by adding two procedures to it, named \textit{Increase Glue Level} and \textit{Bump Glue Variable}, which are called at different states of the search. We denote by $M^{gb}$
the GB extension of the solver $M$.
\begin{table}[h!]
\begin{tabular}{lllllllll}
\multicolumn{8}{c}{\textbf{Algorithm 1: Increase Glue Level}}                                                                                                                                                                                                      & \multicolumn{1}{l}{\;\;\;\;\;\;\;\;\;\;\;\;\;\;\;\;\;\;\textbf{Algorithm 2: Bump Glue Variable}}                                                                                                                                                                                \\
\multicolumn{8}{l}{\textbf{Input:} A newly learned glue clause $g$}                                                                                                                                                                                & \;\;\;\;\;\;\;\;\;\;\;\;\;\;\;\;\;\;\textbf{Input:} A glue variable $v_g$                                                                                                                                                                                      \\
\multicolumn{8}{l}{\begin{tabular}[c]{@{}l@{}}1\;\;\;\;\textbf{For} $i \leftarrow 1 \;to\; \vert g \vert$\\     2 \;\;\;\;\;\;$v_g \leftarrow varAt(g,i)$\\      
3\;\;\;\;\;\;\;$gl(v_g) \leftarrow gl(v_g)+1$\\
4\;\;\;\;\textbf{End}\end{tabular}} & \begin{tabular}[c]{@{}l@{}}
\;\;\;\;\;\;\;\;\;\;\;\;\;\;\;\;\;\;1\;\;$gc({v_g}) \leftarrow \frac{gl(v_g)}{\ \forall_{v_{g'} \in vars(G)} \sum gl(v_{g'})}$\\ 
\;\;\;\;\;\;\;\;\;\;\;\;\;\;\;\;\;\;2\;\;$bf \leftarrow activity(v_g) * gc({v_g})$\\
\;\;\;\;\;\;\;\;\;\;\;\;\;\;\;\;\;\;3\;\;$activity(v_g) \leftarrow activity(v_g) + bf$\\ 
\end{tabular}
\end{tabular}
\end{table}

\smallskip
\noindent \textbf{Increase Glue Level:} Whenever $M^{gb}$ learns a  new glue clause $g$, before making an assignment with the first UIP variable that appears in $g$, it invokes Algorithm 1. For each variable $v_g$ in $g$, its glue level of $v_g$ is increased by 1 (Line 3).  

Clause $g$ is the latest learned clause and all the variables in $g$ are assigned in the current search state. A variable  in $g$ becomes a candidate branching variable again if it becomes unassigned by backtracking. $M^{gb}$ delays the bumping of $v_g$ until it is unassigned by backtracking. The extended solver invokes the bumping routine for $v_g$, right after $v_g$ becomes unassigned by backtracking. 

\smallskip
\noindent \textbf{Bump Glue Variable:}  Algorithm 2 bumps a glue variable $v_g$, which has just been unassigned by backtracking. This is done by adding a \textit{bumping factor } to its activity score (Lines 2, 3). The bumping factor takes two intuitions into considerations: 

\noindent (a) At a given state of the search, the glue centrality of a \glueVar{} measures the glue level of that variable relative to the glue levels of all the \glueVars{}. Given a glue clause $g$, let $\{v_g , v_g'\} \in g$ with $gc(v_g)>gc(v_g')$. Compared to $v_g'$, $v_g$ can potentially affect more blocks of variables and has bigger potential to create conflicts.
\noindent (b) Activity score of a variable measures the extent of involvement of that variable in recent conflicts and indicates its chance to be involved again in a conflict in the near future. 

Heuristically, a good bumping factor combines both of the activity score and glue centrality of the \glueVars{}. Hence, $M^{gb}$ computes the bumping factor for a glue variable $v_g$ by multiplying its current activity score $activity(v_g)$  and its glue centrality $gc({v_g})$. 

\section{Experiments}
We conduct our experiments using \MLDgb{} and \MLDCBTgb{}, which implement the GB method on top of the solvers \MLD{} and \MLDCBT{}, respectively. The branching heuristics in these baseline solvers do not distinguish between glue and nonglue variables.
We run all the 750 main track instances from SAT-2017 and 2018 on 
the same machine as mentioned earlier with 5000 seconds timeout limit for each instance. 

\vspace{-0.1cm}
\begin{table}[]
\caption{Comparisons for the main track instances from SAT-2017, 2018.}
\resizebox{\textwidth}{!}{%
	{\small
\begin{tabular}{|c|c|c|c||c|c|c|||c|c|c||c|c|c|}
\hline
\multirow{3}{*}{\textbf{\begin{tabular}[c]{@{}c@{}}Problems\\ Sets\end{tabular}}} & \multicolumn{12}{c|}{\textbf{Systems}}                                                                                                                                                                                                                                                                                                                                               \\ \cline{2-13} 
                                                                                  & \multicolumn{3}{c||}{\textbf{(A) MLD}}                                                  & \multicolumn{3}{c|||}{\textbf{(B)} $\mathbf{MLD^{gb}}$}                                           & \multicolumn{3}{|c||}{\textbf{(C) MLD\_CBT}}                                                      & \multicolumn{3}{c|}{\textbf{(D)} $\mathbf{MLD\_CBT^{gb}}$}                              \\ \cline{2-13} 
                                                                                  & \textbf{Sat}         & \textbf{Unsat}       & \textbf{PAR-2}                       & \textbf{Sat}                  & \textbf{Unsat}                & \textbf{PAR-2}                      & \textbf{Sat}         & \textbf{Unsat}                & PAR-2                                & \textbf{Sat}                  & \textbf{Unsat}       & \textbf{PAR-2}                       \\ \hline
\multirow{2}{*}{\textbf{SAT-17}}                                                  & \multirow{2}{*}{99}  & \multirow{2}{*}{106} & \multirow{2}{*}{\textbf{1635712.64}} & \multirow{2}{*}{99}           & \multirow{2}{*}{\textbf{107}} & \multirow{2}{*}{1645635.64}         & \multirow{2}{*}{103} & \multirow{2}{*}{\textbf{113}} & \multirow{2}{*}{1565640.80}          & \multirow{2}{*}{\textbf{107}} & \multirow{2}{*}{111} & \multirow{2}{*}{\textbf{1523343.19}} \\
                                                                                  &                      &                      &                                      &                               &                               &                                     &                      &                               &                                      &                               &                      &                                      \\ \hline
\multirow{2}{*}{\textbf{SAT-18}}                                                   & \multirow{2}{*}{136} & \multirow{2}{*}{101} & \multirow{2}{*}{1807074.6}           & \multirow{2}{*}{\textbf{139}} & \multirow{2}{*}{\textbf{102}} & \multirow{2}{*}{\textbf{1770015.2}} & \multirow{2}{*}{\textbf{135}} & \multirow{2}{*}{\textbf{102}}          & \multirow{2}{*}{\textbf{1800012.20}} & \multirow{2}{*}{134}          & \multirow{2}{*}{101} & \multirow{2}{*}{1815868.62}          \\
                                                                                  &                      &                      &                                      &                               &                               &                                     &                      &                               &                                      &                               &                      &                                      \\ \hline
\multirow{2}{*}{\textbf{Overall}}                                                 & 235                  & 207                  & \multirow{2}{*}{3442787.24}          & \textbf{238}                  & \textbf{209}                  & \multirow{2}{*}{\textbf{3415650.6}} & 238                  & \textbf{215}                  & \multirow{2}{*}{3365653}             & \textbf{241}                  & 212                  & \multirow{2}{*}{\textbf{3339211.81}} \\ \cline{2-3} \cline{5-6} \cline{8-9} \cline{11-12}
                                                                                  & \multicolumn{2}{c|}{442}                    &                                      & \multicolumn{2}{c|}{\textbf{447}}                             &                                     & \multicolumn{2}{c|}{453}                             &                                      & \multicolumn{2}{c|}{453}                             &                                      \\ \hline
\end{tabular}}}
\label{table2}
\end{table}

\smallskip
\noindent \textbf{\MLD{} vs. \MLDgb{}:}
Columns (A) and (B) of Table \ref{table2} compare the performance of \MLD{} and \MLDgb{}.   
 Our GB extension achieves a total gain of 5 instances over its baseline, of which 3 are satisfiable and 2 are unsatisfiable. Overall, \MLDgb{} has a smaller PAR-2 score compared to its baseline, which reflects the overall stronger performance of \MLDgb{} over its baseline. 

\smallskip
\noindent \textbf{\MLDCBT{} vs \MLDCBTgb{}:} 
Both solvers solve an equal numbers of instances (Columns C and D, Table \ref{table2}). However, \MLDCBTgb{} solves some of the instances faster than \MLDCBT{}, which is reflected in its overall lower PAR-2 score.   

\begin{figure}[h]
\caption{Solve time comparisons. For any point above 0 in the vertical axis, our extensions solve more instances than their baselines at the time point in the horizontal axis.}
\centering
\label{solveTime}
\includegraphics[width=\textwidth, height=0.50\textwidth]{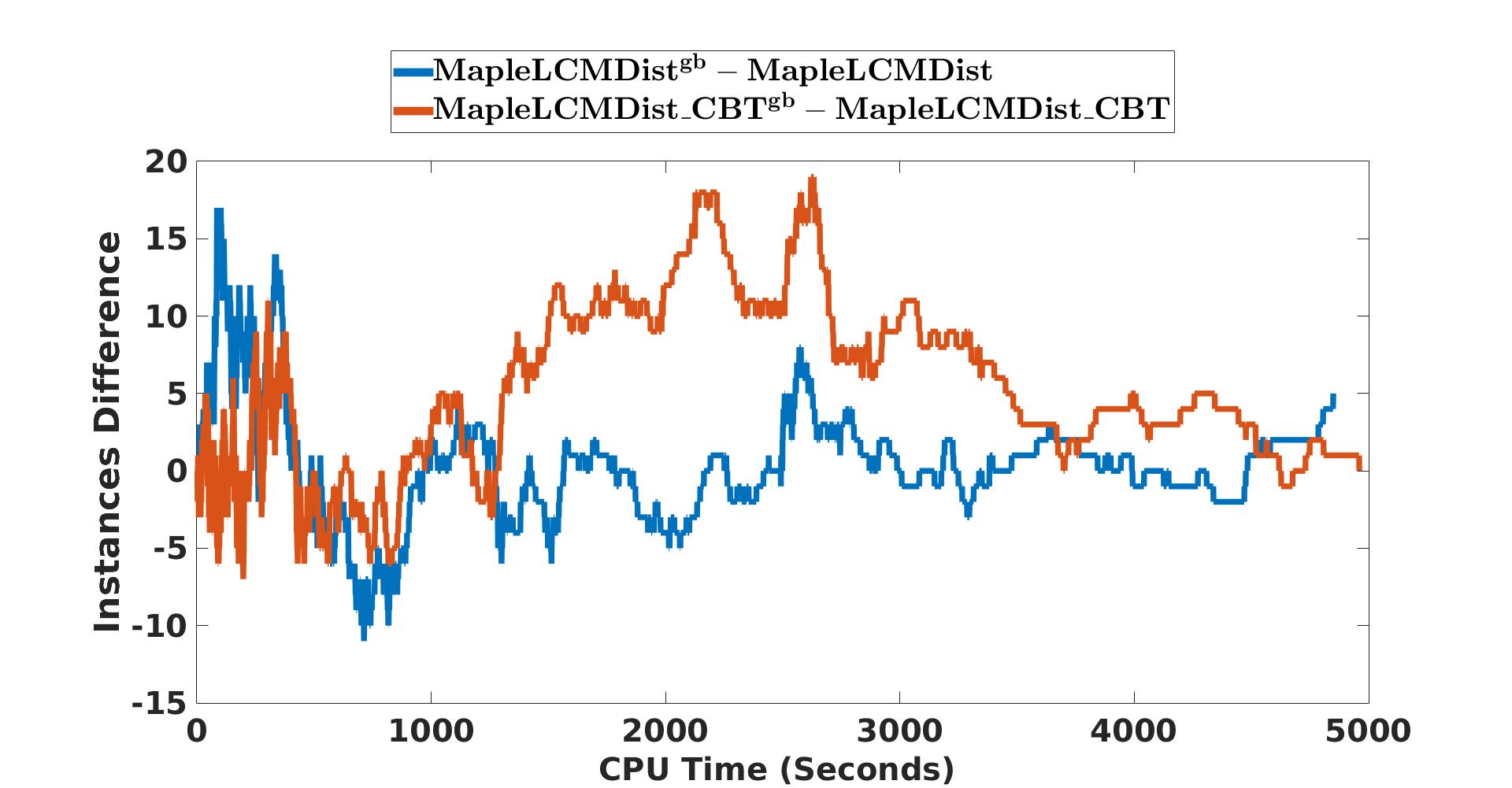}
\end{figure}


\noindent \textbf{Solved Time Comparison:} Figure 3 compares the performance of \MLDgb{} (blue line) and \MLDCBTgb{} (red line) against their baselines, \MLD{} and
\MLDCBT{}, respectively,
This figure plots the difference in number of instances solved as a function of time.
At most points in time, the extended solvers improve on the number of solved problems. This is especially pronounced at first time limits (blue line)  for \MLDgb{}   and at intermediate time  limits (red line) for \MLDCBTgb{}.



Overall, the GB method improves the performance of both baseline solvers, which is reflected in the lower PAR-2 scores of the extended solvers.

\section{Discussion}
In this section we analyze our experimental results reported in the previous section to reveal some insights on the inner workings of the GB method. We limit our discussion with \MLDGBShort{} for the space restraint.

\smallskip
\noindent \textbf{Propagation Rate, Global Learning Rate (GLR) and average LBD:} Table 3 shows the Propagation Rate, Global Learning Rate and average LBD score of the learned clauses for \MLD{} and \MLDgb{} for the 750 instances.
\vspace{-0.1cm}
\begin{table}[!htb]
    \begin{subtable}{.5\linewidth}
      \centering
        \caption{ \textbf{Table 3: Inference and Conflict Efficiency Comparison} }
        \label{TabMetrics}\resizebox{0.9\textwidth}{!}{%
	{\small
        \begin{tabular}{|l|l|l|l|}
\hline
\textbf{System}                                                  & \textbf{PR (A)}                      & \textbf{GLR (B)}                   & \textbf{Average LBD (C)}            \\ \hline
\multirow{2}{*}{\textbf{MapleLCMDist}}                         & \multirow{2}{*}{158.31}          & \multirow{2}{*}{0.40}          & \multirow{2}{*}{27.73}          \\
                                                               &                                  &                                &                                 \\ \hline
\multirow{2}{*}{\textbf{MapleLCMDist}$\mathbf{^{gb}}$} & \multirow{2}{*}{\textbf{160.25}} & \multirow{2}{*}{0.40} & \multirow{2}{*}{\textbf{27.56}} \\
                                                               &                                  &                                &                                 \\ \hline
\end{tabular}}}
    \end{subtable}%
    \begin{subtable}{.5\linewidth}
      \centering
       \label{TabImpactGlueBump} \caption{\;\;\;\textbf{Table 4: Comparison of Glue Decision Measures.} }\resizebox{0.9\textwidth}{!}{%
	{\small
        \begin{tabular}{|c|l|c|c|c|}
\hline
\textbf{System}                                            & \#gd (A)                             & \multicolumn{3}{c|}{\textbf{Average over Glue Decisions}}                                  \\ \hline
\multirow{2}{*}{\textbf{MapleLCMDist}}                     & \multirow{2}{*}{26.8 M}          & \textbf{PR (B) }             & \textbf{LR (C)}                    & \textbf{average LBD (D)}            \\ \cline{3-5} 
                                                           &                                  & \textbf{ 371.21}         & 0.52                           & 23.29                           \\ \hline
\multirow{2}{*}{\textbf{MapleLCMDist}$\mathbf{^{gb}}$} & \multirow{2}{*}{\textbf{28 M}} & \multirow{2}{*}{370.49} & \multirow{2}{*}{0.52} & \multirow{2}{*}{\textbf{22.63}} \\
                                                           &                                  &                         &                                &                                 \\ \hline
\end{tabular}}}
    \end{subtable} 
\end{table}

\MLDgb{} has higher PR (Column A), equal GLR (Column B) and lower average LBD (Column C). Compared to its baseline the bumping of \glueVars{} helps \MLDgb{}  achieve better inference rate, to generate conflicts at an equal rate from which it learns slightly higher quality clauses on average. In \cite{Liang2017}, the authors showed that better branching heuristics increase GLR and lower the average LBD scores of the learned clauses compared to the less efficient ones. Thus our result is largely consistent with the result of \cite{Liang2017}.  

\smallskip
\noindent \textbf{Impact of GB on Glue Decisions:} The GB method is designed to prioritize the selection of \glueVars{}. Here, we take a closer look at the impact of this strategy on glue decisions for \MLDgb{} and its baseline. As shown in Table 4, the GB method in \MLDgb{} indeed increases the average number of glue decisions (Column A) to 28.0 millions (47.85\% of total decisions in \MLDgb{}) from 26.8 millions (47.60\% of total decisions in \MLD{}), which is the average number of glue decisions taken by \MLD{}. 

Though GB method increases total number of propagations (not shown in Table 4) on average, the increased average number of glue decisions lowers the average PR (Column B).  The increased glue decisions also generates conflicts at the same rate (Column C) from which slightly better quality clauses are learned (Column D), on average.


\section{Related Work}
In \glucose{} \cite{AudemardS09}, the authors explicitly increased the activity scores of variables of the learned clause that were propagated by a glue clause. In their work, the bumping was based on VSIDS score bumping scheme. Here, we increase the activity scores of  variables that appear in glue clauses based on glue centrality, which is a structure aware measure, to improve sate-of-the-art CDCL SAT solvers. In \cite{Eigencentrality}, the authors studied the behavior of the CDCL SAT solver \glucose{} w.r.t. \textit{eigencentrality}, a precomputed static measure of the variables in industrial SAT instances. It shows that the branched and propagated variables in \glucose{} has high eigencentrality and the learned clauses contains high eigencentral variables than conflict clauses. In contrast, we dynamically characterize glue and nonglue variables within the course of a search and show that decisions with glue variables are categorically more inference and conflict efficient than decisions with nonglue variables. In \cite{understandingVSIDS}, the authors have shown that the VSIDS heuristic disproportionately branches more often on those variables which are the bridges between communities. Here, we show that CDCL heuristics branches disproportionately more often on glue variables w.r.t. their relatively smaller pool size.  In \cite{betweeness}, the authors exploit the \textit{betweeness centrality} of the variables in industrial SAT formulas to design new heuristics. In their work, betweenness centralities of the variables in a given instance are precomputed before the search begins by using an external python package. In contrast, we compute the glue centrality of the variables during the search without using any external tool. 

\section{Future Work } In this work, we show that variables appearing in glue clauses are more inference and conflict efficient than other variables. We also show that prioritizing glue variable selection improves the performance of CDCL SAT solvers \MLD{} and \MLDCBT{}. In the  future, we plan to pursue the following research questions: (I) are there any correspondence between glue centrality and other centrality measures, such as, eigencentrality, betweenness centrality? (II) In our experiment, some of the benchmark instances have high fraction of glue variables, while for the others only a small fraction of the variables are glue variables.  We plan to study this partition of benchmarks to find out their structural differences.  


\bibliographystyle{plain}
\bibliography{main}
\end{document}